\documentclass[conference]{IEEEtran}
\IEEEoverridecommandlockouts
\usepackage{cite}
\usepackage{amsmath,amssymb,amsfonts}
\usepackage{algorithmic}
\usepackage{graphicx}
\usepackage{textcomp}
\usepackage{xcolor}
\usepackage{booktabs}
\usepackage{array}
\usepackage{comment}
\newcommand{\keyterm}[1]{\emph{#1}}

\def\BibTeX{{\rm B\kern-.05em{\sc i\kern-.025em b}\kern-.08em
    T\kern-.1667em\lower.7ex\hbox{E}\kern-.125emX}}

\begin{document}

\title{RagHarnessDebugger: Jev-Guided Agentic RAG in Contract Inference Repair Auditing}

\author{Moriya Dechtiar$^{*1}$, Daniel Martin Katz$^{2,3,4,5}$, Michael J. Bommarito II$^{5,6}$, Hongming Wang$^{1}$, Sylvain Jaume$^{7}$%
\thanks{$^{*}$Corresponding author: moriya.dechtiar@alumni.harvard.edu}%
\thanks{$^{1}$Harvard University, USA}%
\thanks{$^{2}$Illinois Tech - Chicago Kent College of Law, USA}%
\thanks{$^{3}$CLTDS, Bucerius Law School, Germany}%
\thanks{$^{4}$Yong Pung How School of Law, Singapore Management University}%
\thanks{$^{5}$CodeX - The Stanford Center for Legal Informatics, Stanford University}%
\thanks{$^{6}$Bommarito Consulting, LLC}
\thanks{$^{7}$Massachusetts Institute of Technology, USA}%

}
\maketitle

\begin{abstract}
Agentic systems using retrieval-augmented generation (RAG) must repair their own answers by gathering more evidence, testing alternative policies and deciding if, when, and what repairs are needed. Yet how do we know if a system's repairs are working when metrics like accuracy can be misleading? We present RagHarnessDebugger, a framework that records every repair decision in a Jev-guided harness for contract inference and audits the record to facilitate complete debugging of all decisions taken during a RAG repair journey. RagHarnessDebugger checks that each trace is intact, that each validation applies to the candidate it is counted for, and which answers the acceptance rule can reach. We tested our approach on ContractNLI non-disclosure agreements with two reader models and a replication set of separate contracts. Our debugger's audit layer exposed failure patterns that were hidden under aggregate metrics,  policies acceptance cost of repairs vs regressions and policy acceptance gap of 77\% and 76\% of available but unrealised repairs on development and replication sets. Our audit, together with a labeled error analysis and replay mechanism, allowed following the execution path all the way to the terminal decision - uncovering where missing repairs became lost. Our revised harness improved accuracy over the initial BM25-only baseline by 3.6 points on development contracts and 3.7 points on the replication set, and even improved our full contract access reference score by 1.1 on development and 2.4 on replication set. While our work here demonstrates how the debugger can conduct decision root-cause analysis for missed repairs, the true value lies with providing this debugger for judgment visibility serving continuous improvement of RAG decisions and their explainability.
\end{abstract}


\section{Introduction}
Contracts serve as the foundational documents that support many commercial relationships. Through intricate language, they allocate rights, obligations, risk, and the flow of resources between the parties involved. Both individuals and organizations routinely enter into a litany of agreements, such as customer and supplier contracts, leases, software licenses, employment contracts, and nondisclosure agreements. The creation, review, and interpretation of these documents is time-consuming, expensive, and prone to error~\cite{koreeda2021contractnli, coupette2023law, dechtiar2025smells}.

Automated contract analysis at scale is therefore one of the most practical applications of language models in law. This includes pre-signature drafting tasks such as identifying undefined terms, dangling cross-references and conflicting clauses~\cite{dechtiar2025graph}. The pre-signature stage also includes reviewing drafts produced by counterparties with guidance often provided by negotiation playbooks. Contract analysis is also important in post-signature workflows such as due diligence, where there is often a need to review a large body of executed agreements to flag anomalies such as an unexpected change-of-control trigger or a non-standard limitation of liability~\cite{hendrycks2021cuad}.

Whether it involves a human or machine reviewer, most contract analysis tasks center upon determining the presence, applicability and scope of contractual language. The ContractNLI perspective frames this task as a natural language inference (NLI) problem with evidence identification~\cite{koreeda2021contractnli}.

Clause identification is often an important first step in many contract-related workstreams. However, a clause may only state a general obligation while another clause in the same document could limit the parties to which it applies, add exceptions to its applicability, or otherwise limit the scope of a right or obligation. These nuanced properties of contractual documents complicate the context engineering task. Most notably, in a clause-by-clause characterization of contracts, the key information might be missing, or it could be present but not directly considered by the model~\cite{joren2025sufficient}. Existing benchmarks such as RAGChecker and the Automatic LLMs' Citation Evaluation (ALCE) measure and improve how Large Language Models (LLMs) handle external knowledge and factual accuracy. However, they score retrieval, generation, and citation quality as separate activities~\cite{ru2024ragchecker, gao2023enabling}. At the same time, commercial legal research tools built purely upon retrieval still hallucinate on a substantial share of queries~\cite{magesh2025hallucination}.

Given that AI systems are unlikely to always perform ideally in zero-shot contexts, it is attractive to allow a system to dynamically evaluate its own outputs. For example, leading agentic systems using retrieval-augmented generation (RAG) often try to repair their own answers using a variety of approaches, including gathering more evidence and testing alternative approaches. But the crux of such analysis is the evaluation of the various steps in the RAG process.

In the context of contracts, this might include a structural evaluation to determine whether an evidence package holds every passage a graph-traversal rule requires. It could include a citation check to determine whether a cited passage belongs to that package. While these types of atomistic evaluations provide a measure of comfort, the agentic RAG system still has to choose whether to fetch another passage, remove a distracting one, replace information or simply accept the original results.

A well-designed repair harness makes many small decisions on the path to rendering its final answer. For example, a dynamic harness must determine which passages to retrieve, which unretrieved passage would most improve a package, whether an added passage is a support or a distraction, which proposed edit to apply and how much weight a candidate reading deserves. There are many choices and, in turn, many opportunities for online learning and evaluation.

General-purpose language models could, in theory, be leveraged to conduct evaluation. However, in production and particularly at scale, this can be a time-consuming and expensive proposition. Every LLM call adds latency and cost to a pipeline that might need to process thousands of contracts and components thereof.

Notwithstanding time and cost, the LLM-as-a-judge paradigm has been shown to suffer from position and verbosity biases~\cite{zheng2023judging}. There is also a tendency toward LLM groupthink, whereby an LLM judge drawn from the same model family as the reader tends to share the same blind spots as the model being evaluated~\cite{panickssery2024llm}. Ideally, to mitigate the likelihood of correlated errors, independent assessments that share no inputs or intermediate conclusions should be leveraged in evaluation~\cite{kahneman2021noise}.

Jev is a new general-form classifier from TypeSafe AI, released in September 2026~\cite{typesafe2026jev}, that allows for rapid and accurate evaluation of tasks that can be conceptualized as classification. It is a decision-only model that takes a given hypothesis and candidate passages or edits and returns a ranking or label probabilities~\cite{li2026jev}. Recent work on general tasks shows that Jev can produce more accurate judgments than LLM-as-a-judge at lower cost and latency~\cite{li2026jev}. On ContractNLI, Jev reaches roughly similar aggregate scores to language models~\cite{zhang2026same}.

What matters for a repair harness, however, is less the aggregate score and more the patterns that lie beneath it. At similar overall accuracy, Jev and language models decide different instances differently~\cite{zhang2026same}, which is precisely why detailed debugging is so crucial. In our work, through the audit layer, we found that Jev outperformed the LLM readers only on smaller-scoped decisions, such as ranking evidence, but was less useful as a reader. Jev is inexpensive and fast enough to consult at practically every decision point, so the harness can rank unretrieved passage against ``none of these'' and score every proposed edit on every case, which would be prohibitive with an LLM judge at scale. We therefore build the harness around Jev as its guide. Jev reranks the initial retrieval, ranks residual passages, judges whether an added passage is a distraction, and advises on evidence edits, while an LLM adjudicator makes the final selection among candidate readings. Jev never answers the hypothesis itself, and we train no new model.

A well-guided harness can nonetheless fail silently, and a final accuracy figure does not reveal whether its repairs are working. A harness whose repairs and regressions cancel out looks inert; one whose acceptance rule blocks most of what its adjudicator would have fixed looks merely cautious. Systematic debugging links an observed outcome to the states and decisions that produced it~\cite{zeller2009why}, and in an agentic RAG system that means examining the repair controller alongside the retrieved evidence and the generated answer. A useful candidate can be lost in at least three ways: a validation may not apply to the candidate it is counted for, the acceptance rule may exclude the candidate's label, or no available follow-up action may be able to change the condition that blocks acceptance. When an evaluation reports only accuracy, each of these mechanisms stays hidden, and the gains, the losses, and the cost of repair cannot be told apart.

We present RagHarnessDebugger, a framework for auditing a repair harness for contract question answering. The harness records every retrieval, reading, evidence revision, validation, and adjudication decision, together with the inputs and rules each decision used. An offline auditor then verifies the links between these records, looks for the three failure modes named above, and marks which alternative policies can be replayed exactly from recorded outputs and which would need new model calls. The audit uses no reference labels and makes no model calls. Only afterward do we join the labels, to count repairs (wrong initial answers made right) and regressions (right initial answers made wrong), and to find where missed repairs concentrate. We evaluate two harness versions on matched executions and test whether the mechanisms recur on a disjoint set of contracts. Because every decision is tied to a record, a reviewer can check why the harness changed or kept any answer, which is the form of trust that legal use requires.

The primary contributions of this work are as follows:
\begin{enumerate}
\item RagHarnessDebugger, an open-source framework which features instrumentation for auditable and replayable Jev guided repair, a label-free audit layer, a replay engine bounded by recorded provenance and a complete reference harness.
\item Introducing and demonstrating operational measures that follow a correct candidate through four stages of repair: its existence, its admission under the acceptance rule, its selection, and the capacity of an intervention to change the outcome. Candidate Reachability and Intervention Capability are introduced and are label-free flags producing real predictive runtime value. Terminal paths, missed repairs and policy impact are label dependent and was also shown to provide actionable insights.
\item A verifiable evaluation protocol.  This protocol of integrity measures ensures that every step from model call to reported number can be checked by keying model requests, caching, hash signing artifacts, using an append only ledger, and keeping a hash-linked event chain for each execution. Harness versions can then be paired by execution and verified identical by request key. Before audit begins, all verifications are first validated before any claim, or calculation is done. 
\item Empirical results from two harness versions showing accuracy concealed repair churn, and policy acceptance gap of 77\% and 76\% available yet unrealised repairs. 
A revision informed by trace audits and labelled analysis cut the regression rate from 5.3\% to 0.9\% on development and from 4.4\% to 1.2\% on replication as well as improved accuracy even beyond the whole contract read reference. 
\item Reusable resources: the code, 3,096 audited execution traces, the frozen audit packets, and the labeled evaluations, all will become available at https://github.com/moriyadechtiar/rag-harness-debugger.
\end{enumerate}

\section{Related Work}

\subsection{Contracts, Structure, and Surface Patterns}

In contract tasks, the retrieval of evidence is only an element of the overall work. In the original ContractNLI paper, the authors pair document-level inference with evidence identification while noting that exceptions and negation make the task genuinely challenging~\cite{koreeda2021contractnli}. Other legal benchmarks cover related ground. CUAD, for example, annotates contract clauses for expert review~\cite{hendrycks2021cuad} while LegalBench gathers both legal reasoning and contract-related tasks for evaluating large language models (LLMs)~\cite{guha2023legalbench}.

Much of the related technical work on contracts builds upon or checks against some form of document representation~\cite{singh2025survey}. GRAPH-GRPO-LEX builds contract graphs with ontology-guided extraction and reinforcement learning~\cite{dechtiar2025graph}. Contract smells carry the software idea of code smells over to agreements, where patterns such as redundant language, excess legalese, and overlong clauses can be detected automatically~\cite{dechtiar2025smells}. Disagreement among several LLM judges of an extracted contract graph can also serve as an uncertainty signal that triggers targeted refinement~\cite{dechtiar2026judge}. In sum, contract AI typically reviews surface patterns that can point to deeper issues, and this higher-level approach achieves a degree of success. To obtain greater performance, however, it is arguably important to explore how the evidence beneath the surface can directly inform the way a contract analysis pipeline is refined.

\subsection{Retrieval Control and Feedback}

Adaptive RAG systems decide for themselves when to look for more evidence. Self-RAG trains a model to make that call and to critique its own output with reflection tokens~\cite{asai2023self}. Corrective RAG hands the decision to a retrieval evaluator, which triggers fresh retrieval or knowledge refinement~\cite{yan2024corrective}. VeriHarness separates candidate generation from externally controlled validation, and compares raw diagnostics with feedback that identifies the source of failure, the observed value, and the admissible alternatives~\cite{ray2026structured}. We build upon this work and apply it in the context of contract inference. Our approach applies diagnostic signals to shape the candidates the feedback produces and to decide whether they are validated. It also defines the conditions under which the acceptance rule can return results. 
The harness's main goal is investigating control and retrieval decisions, as well as policy judgments, so it is only natural to use existing formulation of judgment mechanisms and biases that have been studied in humans. This is similar to how~\cite{jones2022capturing} explored cognitive biases to generate testable hypotheses about language-model failures. 
Only, we use the judgment literature differently, as design constraints for the harness (Section~\ref{sec:method}-A). 

\subsection{Sufficiency and Validation}

A second group of systems checks whether the evidence supports an answer before trusting it. Sufficient Context asks whether the retrieved material permits an answer and whether the generator uses it~\cite{joren2025sufficient}. SURE-RAG combines evidence sufficiency with uncertainty to decide when to answer~\cite{qiu2026sure}. PAVE extracts premises and uses support scores to guide revision~\cite{huang2026pave}, and CiteGuard-RAG adds citation constraints, grounding checks, and regeneration~\cite{barua2026citeguard}. Evaluation tools separate the same concerns after the fact. RAGChecker gives fine-grained diagnostics for retrieval and generation~\cite{ru2024ragchecker}, ALCE separates answer correctness from citation quality~\cite{gao2023enabling}, and RAGAs scores RAG pipelines without reference answers~\cite{es2024ragas}. Hu~\cite{oraclegap} measures candidate availability, we add on top of that candidate reachability that separates the availability of a candidate, from the system's ability to reach it with selection, or adjudication. Together these systems show that evidence-sensitive validation is workable (if the validation is reasonably fast and inexpensive). Like RAGAs, our audit needs no labels, but it audits the controller's decisions rather than the answer text. We focus on the narrower question of which evidence changes a bounded repair process proposes, which of them improve the answer and which survive an acceptance check. RagHarnessDebugger adds residual ranking as a complementary signal and records which unretrieved passages (those left out of the initial package) Jev ranked above ``none of these,'' and it audits whether the repairs that followed could change the outcome.

\subsection{When and Where to Revise}

The question of when and where to revise is a key consideration in the context of RAG and adaptive RAG. Self-Refine has one model critique and revise its own output in a loop~\cite{madaan2023selfrefine}, while Reflexion feeds verbal reflections on failed attempts into later trials~\cite{shinn2023reflexion}. Without external feedback, intrinsic self-correction may produce lower accuracy on reasoning tasks~\cite{huang2024large}. Chain-of-Verification reduces reliance on the first answer by answering verification questions separately before the final response~\cite{dhuliawala2024chain}. Return or Revise? weighs revision by its paired repair and harm outcomes and compares it with a draft-free alternative~\cite{motlagh2026return}.
Using LLMs as judges can match human preferences on aggregate but does show position and verbosity biases~\cite{zheng2023judging}, this is one reason our adjudicator sees candidates in a fixed but shuffled order. 
JEV-as-a-Judge uses this decision-only model's label probabilities for direct answering~\cite{li2026jev}, however in our harness, Jev's greatest value is guiding evidence decisions and its advice on answers did not improve performance (Section~\ref{sec:results}).

Agent debugging increasingly works from execution trajectories. Some systems say what failed: MAST derives a taxonomy of multi-agent failure modes from annotated traces~\cite{cemri2025why}, and Who\&When attributes task failures to the responsible agent and the decisive step~\cite{zhang2025which}. AgentRx synthesizes guarded constraints, checks them by program or by model, and produces evidence-backed validation logs for failure attribution~\cite{barke2026agentrx}. Other systems act on the diagnosis. AgentDebug finds root-cause failures and feeds corrective feedback into later attempts~\cite{zhu2025where}, and DoVer tests failure hypotheses by intervening, for example by editing messages or altering plans, and counts a failure as debugged when an intervention resolves it~\cite{ma2025dover}.

Within RAG, Doctor-RAG diagnoses failures and applies localized repairs with prefix reuse~\cite{jiao2026doctor}, and RECTIFY organizes post-evaluation diagnosis, repair proposals, and verification~\cite{murugaraj2026rectify}.

RagHarnessDebugger instead examines the repair controller itself and considers the applicability of each validation to the candidate it supports, the admissibility of recorded conclusions under the acceptance rule and the capacity of follow-up actions to change what blocks acceptance, judged before it runs rather than tested by intervening. 

Its auditor is deterministic and examines correct and incorrect final answers alike. The findings describe the recorded repair mechanism. They do not establish that an answer is semantically correct.

\subsection{Runtime Constraints and Replay}

Work on agent runtimes motivates explicit rules and reproducible records. AgentSpec gives structured rules for enforcing constraints while an agent runs~\cite{wang2026agentspec}, and deterministic replay systems record external interactions and reproduce executions by matching requests~\cite{mudasiru2026replay}. 

Paduraru et al.~\cite{paduraru2026trace} explores a trace-based assurance framework for agentic orchestration instruments executions with step and trace contracts that localize violations, support deterministic replay, and add fault injection.

RagHarnessDebugger uses provenance to identify the boundary of replayed policy comparisons so a replay counts as supported only when the recorded outputs cover every dependency of the alternative decision. We cannot replay if we don't have the complete alternative path routable and rebuilt. A change that needs an unrecorded model response falls outside that boundary. Within supported replays, reference labels count both the regressions a rule prevented and the repairs it blocked. So we are not choosing what is the correct policy and apply it, we report on its impact as the single change in the traced route.

\section{Method: RagHarnessDebugger}\label{sec:method}

\begin{figure*}[htbp]
\centering
\includegraphics[alt={Flow diagram of the harness: retrieval, reading, two gates, repairs, validation, adjudication, and one follow-up branch.}]{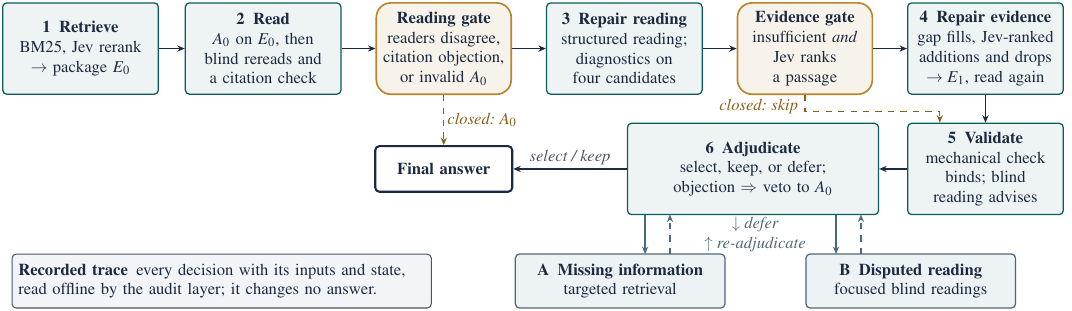}
\caption{The repair harness. Solid arrows are the open path; a closed gate returns the current answer or skips evidence repair (dashed gold), and dashed slate arrows return to adjudication after a follow-up branch. A deferral runs at most one follow-up branch, after which adjudication runs once more; a second deferral keeps $A_0$ and is reported as unresolved. Every decision is recorded for the offline audit (Fig. \ref{fig:audit}).}
\label{fig:harness}
\end{figure*}

RagHarnessDebugger is our system to facilitate both the Jev-guided RAG for contract inference and its audit layer to allow system debugging. The focus is to allow active debugging for contract inference that is guided by Jev and designed by judgment principles discussed below. This is a contract inference RAG pipeline that allows debugging not just its results but all the steps leading to its results. This idea, imported from software engineering, focuses not just on correcting a specific answer, or test case, but also on flagging and correcting harness design decisions that are hindering the system's performance. The key to both layers of repair is the audit layer and active tracing done during the pipeline run. Fig. \ref{fig:harness} shows the harness diagram, its components and flow, and Fig. \ref{fig:audit} shows its audit layer with its captured traces protocol, diagnostic flags and checks.

\subsection{Design Principles}
When we design a system for judgment we can draw inspiration from known mechanisms of human judgment. This topic has been studied a great deal and there are a few principles we thought would be valid to guide the harness judgment. We use these principles as heuristics and look for later validation from actual results. One of these is known as \emph{What you see is all there is}~\cite{kahneman2011thinking} which motivated us to estimate whether what we have is indeed all there is. The risk is our harness not considering there could be other evidence passages that we are missing. We consider the entire residual after $E_0$ is fetched, and a BM25 filter is applied and the top 30 passages (configurable) are sent for Jev ranking with an additional option ``none of these'' and the question: ``Which unretrieved passage would most help decide the hypothesis?'' If any of the 30 passages is ranked higher than ``none of these,'' this flags evidence we missed, which should be included. 
Another concept we incorporated is \emph{Attribute substitution}~\cite{kahneman2002representativeness}. This idea has been used successfully for a variety of tasks when the concept is replacing a hard question with one that we can more easily solve. For instance, instead of asking ``is this label right, given the whole contract?'' we ask four different questions: ``are the citations valid?'', ``do the citations discuss the topic in the hypothesis?'', ``is the evidence closed, that is, is every defined term or dependent party mentioned in the evidence covered by the passages?'', and ``do the selected citations support the label they are matched with?'' So we are measuring structural closure, scoped relevance and answering feasibility, instead of aiming to measure correctness. Another judgment principle is about decomposing a problem and is captured in Kahneman et al.~\cite{kahneman2019structured} as \emph{Mediating assessments}. The idea is when assessment is required, to break it down to parts in order to be able to decrease the scope of what needs to be assessed. Humans are less likely to make mistakes when assessing smaller scopes. This principle was used to construct the structured reading, in which we ask for the answer in a specific order of steps, and to enrich the adjudication information with a set of smaller judgment calls on specific parts of the decision. For instance: is this actor or action supported by the evidence, versus is this entire hypothesis supported? This is helpful, for instance, to eliminate specific labels: a hypothesis cannot be supported by the evidence if none of its components are mentioned in it. To encourage objective judgment and reading, we also leverage three more principles: \emph{Anchoring}~\cite{tversky1974judgment}, where we prevent a reader from seeing the earlier answer of a previous read, \emph{Independent judgment}~\cite{kahneman2021noise} which is implemented by blinding a provider to the second provider's judgment, and \emph{Considering the opposite}~\cite{lord1984considering}, where we create a counterexample challenge in adjudication of a disputed proposition.

\subsection{Stages}

The harness runs through six stages, plus follow-up loops. The pipeline can run several follow-up loops; for this paper it ran with a single follow-up loop, because of resources.

\emph{1) Retrieval.} This stage's input is the hypothesis and a set of passages of a complete contract. It applies a BM25~\cite{robertson2009bm25} (rank\_bm25 0.2.2; $k_1$=1.5;  b=0.75; $\epsilon$=0.25) filter to the contract's passages against the hypothesis. Both passages and hypothesis were lowercased and tokenized with [a-z0-9]+, but no stemming, stopword removal or query expansion was done. It then uses Jev to rerank the fetched passages, and uses the top passages to fill 800 proxy tokens (rendered characters divided by four) of our initial evidence package $E_0$. We use whole passages in source order.

\emph{2) Reading and first checks.} This stage takes $E_0$ and the hypothesis and produces the first provider's answer $A_0$ on $E_0$, with citations. We then run two blind readings, one by the same reader and one by the other provider, with a provided rubric explaining the labels: ``Entailment means the passages establish the hypothesis. Contradiction means they establish its negation. NotMentioned means they are neutral to the claim: discussing the topic is not enough to entail or contradict it. If you suspect the passages are incomplete rather than the contract being silent, begin the rationale with INSUFFICIENT\_PACKAGE.'' We also provide $E_0$ and the hypothesis. 
In addition to the blind readings, we perform a citation check only examining the cited passages in $A_0$. A \keyterm{reading gate} opens if the readers disagree with $A_0$, if the citation check objects, or if $A_0$ is invalid. Otherwise we return $A_0$. Since this is a terminal path, we treat the two blind readings with this gate as a basic reference policy and name it \keyterm{guarded rereading}. If the gate opens, we continue to possible reading repairs. At this point the evidence pack continues to the next stage.

\emph{3) Reading repair.} The input for this stage is the readings from stage 2 with their answers, $E_0$ and the checks done. We perform a third reading here which is the \keyterm{structured reading}: ``Decide the hypothesis in four steps, written compactly in the rationale: support; contradiction or qualification; parties and conditions; label.''
We also run checks to obtain more diagnostics: a citation check on new citations; a counterexample challenge to $A_0$, for both the plain and structured readings; the \keyterm{inference check}, which marks each element of the hypothesis as supported, contradicted, or unresolved, with quotes verified in code, and informs only the adjudicator; Jev's verdict on $A_0$ for both readings; structural gaps based on the contract graph; and Jev's residual ranking. All of these, with the four candidates on $E_0$, are considered for evidence interventions. The \keyterm{evidence gate} opens only when a reading of $E_0$ reports that the package is insufficient \emph{and} Jev ranks some unretrieved passage in the residual above ``none of these.''
If no evidence intervention is needed, we skip to stage 5 for validation. 

\emph{4) Evidence repair.} If a structural gap has been identified, or some residual passages are ranked above ``none of these,'' the code builds new passages and creates $E_1$. Since we are filling our evidence pack to a fixed budget, we may also need to drop passages to make room for new ones, and use Jev to do so. The result, $E_1$, receives the same readings as the initial evidence package.

\emph{5) Validation.} In this stage we validate our candidates, making sure to check a candidate only on the evidence pack it used. Candidates that fail the mechanical checks are dropped. We then recover the blind reading result for the valid candidates and attach them as support or objection to be adjudicated in stage 6. 

\emph{6) Adjudication.} An adjudicator chooses among the valid candidates, with $A_0$ among them, unlabeled. It can select a candidate, keep the current answer, or defer with a reason. Before any adjudication call, a decision rule runs in code: with no valid candidate, the harness keeps $A_0$; if every valid candidate gives the same label, each supported and none objected to, it returns that label. Otherwise the adjudicator decides. The \keyterm{veto} sends back to $A_0$ any selected candidate, other than $A_0$, that carries a blind-reader objection. The selection itself stays in the record, so we can also report the answer the harness would have given without the veto.

A deferral triggers one follow-up branch, never both. For missing information, the harness runs a targeted retrieval for the named proposition and passes the result, possibly empty, onward as evidence. For a disputed interpretation, both readers give a focused blind reading of the proposition. Adjudication then runs once more. After the run completes, the auditor examines each case's recorded trace offline. It changes no answer and no decision.

\begin{figure}[t]
\centering
\includegraphics[alt={Audit layer flow: frozen records, provenance verification, component checks, typed findings, freeze; labels join only afterwards.}]{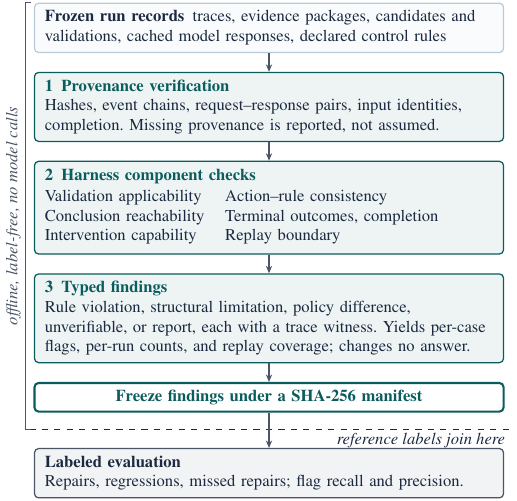}
\caption{The audit layer. Everything above the dashed line runs offline, without reference labels or model calls. Labels join only after the findings are frozen.}
\label{fig:audit}
\end{figure}

\subsection{Implementation}\label{sec:impl}
The system is implemented as an open-source pipeline written in Python. The harness code, prompts, evaluation code, and per-case records are released with the paper. The reader models are gpt-5.6-terra (reasoning set to ``none'') and gemini-3.5-flash-lite (``minimal''). These were selected because they allow us to turn reasoning off and evaluate the harness features in isolation. Temperature was also set to zero for the same reason. 
For ranking and selection of options we use Jev (typesafe/jev-1.13). Packages are capped at 800 proxy tokens for the initial evidence package, plus up to 600 for edits. Frozen append-only ledgers, traces and code packages are kept and protected with hash signatures.

\subsection{Instrumentation Contract}\label{sec:contract}

To allow debugging, we must audit every repair-related decision with the information leading up to it. Therefore, RagHarnessDebugger requires four kinds of record: every model request is keyed by a hash of its exact payload and bound to its cached response; every candidate is stored with its full evidence package, its citations, its mechanical validity, and the readings counted for it; the adjudicator's selection is recorded, with its complete package, before any gated action; and each case keeps a hash-linked event chain from initial retrieval to final answer. Together these records make every path and component of the harness auditable, and every case replayable for what-if analysis.
This also protects us from model changes or drift that could alter our harness analysis outcomes.
\subsection{The Audit Layer}\label{sec:auditor}
The audit layer is what makes debugging possible. When the harness completes its six stages on a test case or a test suite, the audit layer runs offline on the frozen artifacts that the instrumentation contract requires. In a software integrated development environment, the debugger is part of the execution, and the engineer can stop at any time to inspect the current state and check its validity. Our audit layer simulates this with captures that allow complete replays: capturing everything allows rerunning the same flow with changes, using only what was captured at run time. It uses no reference labels and makes no model calls. Its components (Fig. \ref{fig:audit}) fall into three stages.

\begin{enumerate}
\item \emph{Provenance verification.} Before we can trust our analysis, we validate the traces: hash checks on artifacts, event chains, request-response pairs, input identities, execution completion, and Jev's legacy receipts where they exist. Missing provenance is reported, not assumed.

\item \emph{Harness component checks.} Once the artifacts are confirmed, the analysis can begin. Record routing rebuilds the harness's decisions from its signals, so that repair mechanisms and alternative policies can be examined and replayed. Six checks run on the rebuilt decisions.
\emph{Validation applicability}: did the validator read this candidate's package and hypothesis? Did validation pass? 
\emph{Conclusion reachability}: which answers are present at a given stage, and which can be accepted or selected? 
\emph{Intervention capability}: can this intervention change the set of candidates eligible for acceptance, or is its impact limited?
\emph{Action--rule consistency}: do the recorded actions follow the declared rules, and are formatting violations kept from invalidating answers?
\emph{Terminal outcomes and completion integrity}: did the execution complete, and did each repair produce a valid, schema-compliant result?
\emph{Replay boundary}: which counterfactual replays do the recorded dependencies support, and does the tracing protocol need updates to cover more replays? This matters most when harness changes break backward compatibility.

\item \emph{Typed findings.} Each finding has one type (rule violation, structural limitation, policy difference, unverifiable, or report), each with the trace records that witness it. Both the findings and the replay records are frozen under a SHA-256 manifest. They are then used for analysis across versions, runs, injected defects, and specific components. A separate evaluator joins the reference labels to measure repair outcomes, triage quality, and policy trade-offs.
\end{enumerate}

\section{Experimental Setup}\label{sec:setup}

\subsection{Data}\label{sec:data}

Cases come from ContractNLI~\cite{koreeda2021contractnli}, a dataset of hypotheses about non-disclosure agreements. Each hypothesis is a question about the contract with three possible labels: the contract supports it, contradicts it, or does not mention it. The dataset also marks evidence spans, the exact locations in the contract where the answer is supported or contradicted.

Each case runs once with each of two reader models, so every case yields two \keyterm{executions}. This keeps the evaluation from being coupled to one model and its biases, and the second model makes the counterexample checks, the guarded rereading, and the support or objection for adjudication possible.
Of the 123 contract documents in the ContractNLI test split, we used 100 for development and 22 for replication; one was a duplicate.
Each contract has 17 standard hypotheses. We sampled 4 at random from each development contract, giving 400 cases and 800 executions.
Development served as our training data in the sense that we identified root issues in the harness from its runs and audit findings. To test whether the debugging was effective, we set aside the 22 replication contracts and used all 17 hypotheses for each, giving 374 cases and 748 executions. Replication data was used only for reporting and never to make harness changes: it tests whether a fix made on development worked.
Its initial evidence packages and answers had been produced and inspected in an earlier evaluation, so it is a replication on disjoint contracts, not an untouched hold-out set.

\subsection{What We Ran}
RagHarnessDebugger aims to debug not only a specific repair but the entire repair harness, because a repair depends on the whole RAG flow and its decisions, checks, and mechanisms. We therefore compare two versions of our Jev-guided RAG repair harness: its pre-debug (first) version and its post-debug (revised) version.

After the run on the development cohort, we ran the audit layer. Without gold labels, the action--rule consistency check found deferrals that were triggered but not acted on, because they relied on information that did not exist in the deferred cases. The same check found selections that matched a rule but were rejected in adjudication because of a parsing defect. The labeled pass then exposed further design issues: regression--repair trade-offs, and a gate policy that was too loose and logically incorrect. The first version let the inference check's element flags also open a repair gate: the hypothesis was decomposed into five elements, and the gate opened if any one was flagged. This wrongly required every element to carry the whole hypothesis's label: for a hypothesis to be not mentioned, not every component of it must also be not mentioned. The first version also allowed evidence additions and drops whenever the Jev score passed a fixed threshold, which did not change accuracy but did shift the balance between repairs and regressions. Fixing these issues (the gating policy and the repair--regression trade-off) produced the revised version.
The data, models, and prompts are the same in both versions; the policies, gating, and component design changed to correct the issues found.

\subsection{What We Compare}
To measure whether debugging worked, we compare both final outcomes and the behavior of the entire harness. Complex systems often have changes with ripple effects that both help and harm, much as a software bug fix can introduce new defects. We compare the 1,548 executions across both cohorts in pairs, matched by case and version identifier. Paired executions match on the initial package $E_0$, the initial answer $A_0$, its citations, and the recorded initial-answer request key, so differences come from the repair harness's policies and mechanisms.

Our baselines are the reader's answer on a BM25-only package and $A_0$ on the Jev-guided package; we expect the harness to improve on both. A reading of the complete contract serves as a reference only: it is not realistic for long contracts. The auditor's analysis of the harness also identified a cheap alternative, the \keyterm{guarded rereading}: the model-agreement policy behind the reading gate, with no repairs applied.
We compare accuracy across all of these. Because cases from the same contract are not independent, we compute unadjusted 95\% percentile intervals from 2,000 bootstrap resamples of contracts, keeping every sampled hypothesis and both reader executions within each contract. Accuracy differences are paired by execution.

Only harness runs show repair behavior. For them we measure the \keyterm{repair rate}, the share of initial errors corrected, and the \keyterm{regression rate}, the share of initially correct answers made wrong. An \keyterm{available repair} is an initial error for which at least one mechanically valid candidate carries the reference label; it is \keyterm{missed} if the final answer is still wrong. The recorded provenance and event chains locate each missed repair by its terminal path, where in the repair flow it was lost; the path does not assign causal responsibility.
We measure policy effects by provenance-verified replay of the revised runs. For the veto, a regression is \emph{prevented} when the recorded final answer is correct but the veto-off answer is wrong, and a repair is \emph{blocked} in the reverse case. Replay identifies the actions and signals; measuring their repair and regression effect uses gold labels.

The auditor runs on all frozen, hash-matched artifacts first, and its findings are frozen before any gold-label metric is computed. For each label-free flag we report the share of executions flagged, the share of missed repairs captured (recall), and the share of flagged executions that contain a missed repair (precision), with the same contract-cluster bootstrap. The reference selections are the reading-gate selection, the veto selection, and the unresolved cases; uniform random review gives the expected recall, and a matched comparison cuts the reading gate's selection to the blocked-conclusion flag's review budget. Costs are settled API costs per 1,000 executions, deduplicated by request and including both initial readers; reused requests keep their originally recorded costs.

\section{Results}\label{sec:results}

\subsection{Accuracy Hid Repair Churn}

As shown in table \ref{tab:churn}, the first version achieved the same accuracy as the initial answers, despite making both correct and incorrect changes. The repairs were effectively canceled out by regressions, a phenomenon we call \keyterm{repair churn}. This works in two steps. The audit helped identify these regressions, which we then addressed in the revised version. On identical inputs, the regression rate dropped from 5.3\% to 0.9\% on development and from 4.4\% to 1.2\% on replication. The repair rate also decreased, from 20.9\% to 9.5\% on development and from 16.6\% to 9.3\% on replication. Nevertheless, the revised version's paired accuracy gain over the first version was positive in both cohorts (+1.25 and +1.07 points, though both intervals include zero), while its cost per 1,000 executions fell by 30.5\% on development and 32.3\% on replication.

\begin{table}[htbp]
\caption{Accuracy by condition, and change from $A_0$ with 95\% contract-cluster bootstrap intervals (both readers pooled)}
\label{tab:churn}
\centering
\setlength{\tabcolsep}{3pt}
\begin{tabular}{@{}lrlcc@{}}
\toprule
Condition & Acc.\ (\%) & $\Delta$ pp [95\% CI] & Repair & Regr. \\
\midrule
\multicolumn{5}{@{}l}{\emph{Development (800 executions)}} \\
BM25 only & 77.75 & $-$2.50 & --- & --- \\
Jev-guided, $A_0$ & 80.25 & & --- & --- \\
Guarded rereading & 80.50 & $+$0.25 [$-$0.62, 1.12] & 5.7\% & 1.1\% \\
First version & 80.12 & $-$0.12 [$-$2.00, 1.63] & 20.9\% & 5.3\% \\
Revised version & \textbf{81.38} & $+$1.12 [0.12, 2.25] & 9.5\% & 0.9\% \\
\quad Revised $-$ first & & $+$1.25 [$-$0.50, 3.25] & & \\
Whole contract & 80.25 & \phantom{$+$}0.00 & --- & --- \\
\midrule
\multicolumn{5}{@{}l}{\emph{Replication (748 executions)}} \\
BM25 only & 77.01 & $-$2.81 & --- & --- \\
Jev-guided, $A_0$ & 79.81 & & --- & --- \\
Guarded rereading & 80.48 & $+$0.67 [$-$0.53, 1.87] & 9.3\% & 1.5\% \\
First version & 79.68 & $-$0.13 [$-$1.74, 1.47] & 16.6\% & 4.4\% \\
Revised version & \textbf{80.75} & $+$0.94 [0.00, 2.01] & 9.3\% & 1.2\% \\
\quad Revised $-$ first & & $+$1.07 [$-$0.27, 2.41] & & \\
Whole contract & 78.34 & $-$1.47 & --- & --- \\
\bottomrule
\end{tabular}
\end{table}

\begin{figure}[htbp]
\centering
\includegraphics[alt={Stacked bars: terminal path of every initial error, per cohort; most available repairs were missed, mainly by the veto.}]{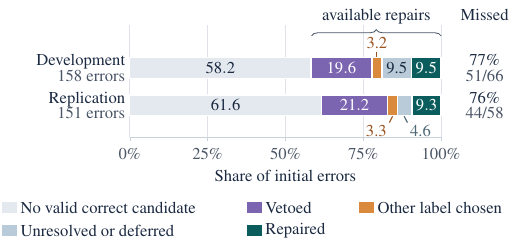}
\caption{Where each initial error ended, revised harness (percent of initial errors). The brace marks errors with a valid correct candidate (available repairs), drawn on the development bar; the right column gives the share and count of those that were missed.}
\label{fig:terminal}
\end{figure}

\subsection{Acceptance Discarded Most Available Repairs}

Fig. \ref{fig:terminal} shows that no valid candidate carried the correct label for most initial errors, making them unreachable.
We observe that when a valid candidate carried the correct label, the harness rarely made use of it.
Indeed, the harness missed 77\% of available repairs on development and 76\% on replication. Furthermore, it missed none of these when the reading gate was closed.
Finally, we observe that both candidate generation and acceptance limit the repair
and that acceptance discards most of what generation found.

\begin{figure}[t]
\centering
\includegraphics[alt={Missed repairs captured versus executions flagged, five label-free flags, two cohorts; all lie well above random review.}]{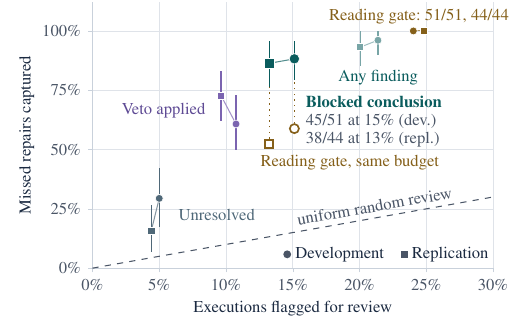}
\caption{Label-free flags scored against missed repairs. Circle: development (51 missed repairs, 800 executions); square: replication (44, 748). Bars are 95\% contract-cluster bootstrap intervals. Hollow markers: the reading gate's selection cut to the blocked-conclusion flag's review budget (30/51 and 23/44 captured; point estimates). The dashed line is the expected recall of uniform random review. ``Any finding'' counts every finding type except report.}
\label{fig:flags}
\end{figure}

\subsection{Label-Free Flags Concentrated Missed Repairs}

The label-free flags identified a large share of missed repairs (Fig. \ref{fig:flags}). In particular, the blocked-conclusion flag captured 88\% of missed repairs on development and 86\% on replication, while selecting only 15\% and 13\% of executions, respectively. Under the same review budget, it identified 1.5 times as many missed repairs as the reading gate's selection on development (45 vs. 30) and 1.65 times on replication (38 vs. 23). Note that these flags identify executions worth reviewing, rather than incorrect answers. In fact, most flagged executions produced correct answers. Moreover, the flags also maintained their recall on the replication contracts, which were not used to develop the rules.

\subsection{Replay Shows What the Veto Costs}

Replaying the revised runs with the veto disabled reveals its benefits and costs. The veto prevented 87\% of potential regressions on development (40 of 46) and 80\% on replication (28 of 35), but also blocked 63\% (25 of 40) and 59\% (20 of 34) of potential repairs, respectively. Its net effect was positive in both cohorts, although at the cost of many missed repairs. However, the terminal paths in Fig. \ref{fig:terminal} show that a missed repair on the veto path does not necessarily mean that removing the veto would have led to a correct answer.

\subsection{Follow-Ups Ran but Could Not Change Answers}

We observe that none of the 33 follow-ups on development changed the final label. The audit helps explain why by examining what each follow-up could have changed, without relying on ground-truth labels. Each follow-up had a predefined set of reachable labels, and focused reading could only select among labels already present in its candidates. Across all 63 follow-ups, 36 focused readings involved initially incorrect answers. In 27 of these cases, the correct label was outside the reachable set, making repair impossible regardless of which candidate was selected. Thus, a successfully executed follow-up does not necessarily mean that a repair was possible.

\section{Discussion and Future Work}
\emph{Where repair fails.} RagHarnessDebugger is a trace-based method to identify where agentic repair fails. These failures can come about from multiple sources, including during candidate generation, validation applicability, acceptance, intervention, or the recording of dependencies needed to replay a policy. It can be important to understand where these failures arise. An unchanged answer, for example, may indicate that no alternative was generated, that a viable candidate was blocked, that a follow-up could not introduce a new candidate, or that retaining the initial answer was the correct decision. Even when repairs and regressions cancel each other out, the regression rate has particular importance in legal applications because it measures how often a system overturns an answer that was already correct.

\emph{Acceptance is a policy choice with a measurable cost.} A veto can transform a net-negative adjudicator into one that is net-positive by blocking most of the repairs the adjudicator would otherwise have accepted.  The question is not whether to accept more or less, but to quantify the benefits and costs of an acceptance rule before deployment.  Recorded traces make this inexpensive, since when downstream requests are cached, alternative acceptance rules can be replayed exactly over past executions without additional inference.  This allows policy comparisons to assess both the repairs gained and the regressions introduced, rather than optimizing acceptance in isolation.

\emph{Flags are a triage instrument.} A flag that puts 88\% of missed repairs into 15\% of executions (86\% into 13\% on replication) lets a reviewer quickly check where repair broke down, without knowing in advance which answers are wrong.  It does not identify wrong answers, and it is not meant to.   The flags are therefore a triage instrument.  Their value lies in directing limited review effort towards potential failures, rather than replacing the reference labels needed to establish correctness. 

\emph{Provenance makes the resulting decisions auditable.}  The recorded trace gives direct answers as to why an answer changed or persisted, which rule blocked a candidate, and what a follow-up intervention could have changed. In legal settings this is important because both the correctness of an answer and the justification for changing or retaining it matter. 

The results motivate targeted interventions at distinct points in the repair process. When a focused review identifies a likely candidate-generation failure, the harness could generate and validate a new candidate for the original hypothesis, allowing an alternative absent from the initial candidate set to be considered. When an acceptance rule appears overly restrictive, the system could seek additional independent support before relaxing the rule. In both cases, candidate-generation changes and acceptance policies should be evaluated separately, with alternative acceptance rules tested through exact replay before deployment. 

Finally, deploying the auditor at execution time and evaluating it across additional legal tasks, contract types, and models would help distinguish general properties of agentic repair from findings specific to the current harness and experimental setting. 

\section{Conclusion}
Agentic systems using retrieval-augmented generation (RAG) must repair their own answers by gathering more evidence, testing alternative policies and deciding if, when, and what repairs are needed.  Yet how do we know if a system’s repairs are working when metrics like accuracy can be misleading?  RagHarnessDebugger makes it possible to review agentic repair as a sequence of decisions rather than a single outcome. By recording these events in traces, RagHarnessDebugger provides a basis for evaluating repair behavior, comparing policies, and auditing interventions beyond what aggregate accuracy alone can reveal. This work illustrates the impact of debugging as a tool to iteratively improve systems, and as systems improve themselves. 

\IEEEtriggeratref{31}

\end{document}